\documentclass[reqno]{amsart}
\usepackage{mathtools,amssymb}
\usepackage{color}

\def\redcol{black}

\def\R{{\mathbb R}}
\def\Z{{\mathbb Z}}

\def\ux{{\underline{x}}}

\def\uy{{\underline{y}}}

\def\cC{{\mathcal C}}

\def\cF{{\mathcal F}}

\def\cL{{\mathcal L}}
\def\cM{{\mathcal M}}
\def\cN{{\mathcal N}}
\def\cP{{\mathcal P}}

\def\cV{{\mathcal V}}

\def\dim{{\rm dim}}

\def\grad{{\rm grad}}

\def\rank{{\rm rank}}
\def\Pen{{\rm Pen}}

\def\1{{\bf 1}}
 
\def\uV{{\underline V}}

\def\Z{\theta}
\def\uZ{\underline{\Z}}

\def\argmin{{\rm argmin}}
\def\ran{{\rm range}}
\def\X0{X_0}

\def\eqnn{\begin{eqnarray*}}
\def\eeqnn{\end{eqnarray*}}
\def\eqn{\begin{eqnarray}}
\def\eeqn{\end{eqnarray}}

\def\prf{\begin{proof}}
\def\endprf{\end{proof}}

\theoremstyle{plain}
\newtheorem{theorem}{Theorem}[section]

\newtheorem{proposition}[theorem]{Proposition}

\newtheorem{lemma}[theorem]{Lemma}

\newtheorem{remark}[theorem]{Remark}

\numberwithin{equation}{section}

\begin{document} 

\title[Geometrically adapted gradient descent in DL]
{Global $\cL^2$ minimization at uniform exponential rate via geometrically adapted gradient descent in  Deep Learning}

\author{Thomas Chen}
\address[T. Chen]{Department of Mathematics, University of Texas at Austin, Austin TX 78712, USA}
\email{tc@math.utexas.edu}

\begin{abstract}
We consider the scenario of supervised learning in Deep Learning (DL) networks, and exploit the arbitrariness of choice in the Riemannian metric relative to which the gradient descent flow can be defined (a general fact of differential geometry).   
In the standard approach to DL, the gradient flow  on the space of parameters (weights and biases) is defined with respect to the Euclidean metric. Here instead, we choose the gradient flow with respect to the Euclidean metric in the output layer of the DL network. This naturally induces two modified versions of the gradient descent flow in the parameter space, one adapted for the overparametrized setting, and the other for the underparametrized setting. 
In the overparametrized case, we prove that, provided that a rank condition holds, all orbits of the modified gradient descent drive the $\cL^2$ cost to its global minimum at a uniform exponential convergence rate; one thereby obtains an a priori stopping time for any prescribed proximity to the global minimum. We point out relations of the latter to sub-Riemannian geometry. Moreover, we generalize the above framework to the situation in which the rank condition does not hold; in particular, we show that local equilibria can only exist if a rank loss occurs, and that generically, they are not isolated points, but elements of a critical submanifold of parameter space. 
\\ \\
MSC classes:	57R70, 62M45
\end{abstract}

\maketitle

\section{Introduction and Main Results}

Use of the gradient descent flow generated by the cost (loss) function is ubiquitous as a method for cost minimization in the training of Deep Learning (DL) networks. While time proven and very successful in general, its implementation exhibits inherent challenges, including the trapping of orbits by approximate or exact local cost extrema, as a consequence of the highly complex cost landscape. Efficient algorithms have been devised to compensate for these obstacles in various scenarios, but from a mathematically rigorous perspective, there is at present insufficient insight into the conceptual and geometric core of the problem.

Our main goal here is to develop a rigorous mathematical understanding of some crucial dynamical systems aspects of supervised learning in DL networks; we do not focus on the development of efficient algorithms, or exploration of specific application cases in the current paper. The definition of a gradient vector field necessitates the {\em  choice} of a Riemannian metric structure, a general fact of differential geometry. In the standard approach to DL, the gradient of the cost function is determined using the {\em Euclidean metric on the space of parameters} (weights and biases); this choice is ad hoc, and not motivated by geometric reasons. As explained in \eqref{eq-ux-graddesc-4-0-0}, below, this {\em defines a highly nontrivial flow in the output layer of the DL network determined by the Neural Tangent Kernel (NTK), \cite{jacgabhon}}; the latter is the reason for the trapping of orbits.

We exploit the arbitrariness of choice in the Riemannian metric relative to which the gradient flow can be defined, both on the side of the parameter space, and in the output layer of the DL network. {\em Choosing the gradient flow with respect to the Euclidean metric in the output layer} of the DL network (instead of picking it on the space of parameters), we introduce two modified versions of the gradient descent flow in the parameter space (by using a preconditioner), one adapted for the overparametrized setting, and the other for the underparametrized setting. Both have a natural invariant geometric meaning, in terms of the pullback vector bundle structure in the overparametrized, and the pushforward vector bundle structure in the underparametrized setting. In the output layer, the Euclidean structure greatly simplifies the cost landscape, and trivializes the gradient flow.

In the overparametrized case, we prove in Theorem \ref{thm-overparam-1-0} that, provided a rank condition holds, all orbits of the modified gradient flow drive the $\cL^2$ cost to its global minimum at a uniform exponential convergence rate; this yields an a priori stopping time for any prescribed proximity to the global $\cL^2$ cost minimum. In Section \ref{ssec-subRiem-1-0}, we point out relations of the latter to sub-Riemannian geometry. 
{\color{\redcol} It is known that invertibility of the NTK implies that zero loss is reachable, \cite{lidingsun,zha}. However, trapping of orbits by approximate local minima of the standard gradient descent flow can occur; this is addressed } 
in Section \ref{ssec-StdGradFlw-1-0}. In Section \ref{sec-rankloss-1-0}, we generalize the above framework to the situation in which the rank condition does not hold; in particular, we show that local equilibria can only exist if a rank loss occurs, and that generically, they are not isolated points, but elements of a critical submanifold of parameter space; 
{\color{\redcol} related results for global minima of overparametrized networks were established in \cite{coo}.} 
In the underparametrized situation, we show in Theorem \ref{thm-underparam-1-0} that the analogous version of the modified gradient flow is mapped to a geometrically natural constrained gradient flow. In Section \ref{sec-borderline-1-0}, we show that in the borderline case separating the strictly overparametrized and underparametrized situations, the two modified gradient flows coincide. A key objective in this work is to elucidate and exploit the link between gradient flows in the parameter space and in the output layer.

In our works \cite{cheewa-2,cheewa-2-2} with P. Mu\~{n}oz Ewald, we provided an explicit construction of globally minimizing weights and biases for the $\cL^2$ cost in underparametrized ReLU DL networks. In \cite{cheewa-3}, we proved that those global minimizers can generically not be approximated via the gradient descent flow algorithm; this fact remains valid for the modified gradient descent flow obtained in this paper for the underparametrized case. Some key aspects of the analysis presented in the work at hand are related to the approach in \cite{che-1} to classical Hamiltonian systems with non-holonomic constraints; see also \cite{arnnov-1,belris-1} for related background on sub-Riemannian geometry, and \cite{we-1,ehanli-1} on dynamical systems and control theory aspects of machine learning. We refer to \cite{arora-ICM-1,hanrol,grokut-1,lcbh,nonreeste} and the references therein for some contextually related background on neural networks.

\subsection{Definition of the mathematical model}

We consider a DL network in the context of supervised learning, with training inputs given by $x_{j}^{(0)}\in \R^M$, $j=1,\dots,N$ where we refer to $\R^M$ as the input space.
We assume that the outputs are given by $y_{\ell}\in\R^Q$, $\ell=1,\dots,Q$. We introduce the map 
\eqn
	\omega:\{1,\dots,N\}\rightarrow\{1,\dots,Q\}\,,
\eeqn 
which assigns the output label $\omega(j)$ to the $j$-th input label; that is, $x_{j}^{(0)}$ corresponds to the output $y_{\omega(j)}$. We write 
\eqn
	\uy_\omega:=(y_{\omega(1)}^T,\dots,y_{\omega(N)}^T)^T\in\R^{NQ}
\eeqn 
where $A^T$ is the transpose of the matrix $A$. We denote by $N_i$ the number of training inputs belonging to the output vector $y_i$, $i=1,\dots,Q$. 

We assume that the DL network contains $L$ hidden layers; the $\ell$-th layer is defined on $\R^{M_\ell}$, and recursively determined by
\eqn
	x_j^{(\ell)} = \sigma(W_\ell x_j^{(\ell-1)} + b_\ell) \;\;\in\R^{M_\ell}
\eeqn
via the weight matrix $W_\ell\in\R^{M_\ell\times M_{\ell-1}}$, bias vector $b_\ell\in\R^{M_\ell}$, and activation function $\sigma$. We assume that the output layer 
\eqn
	x_j^{(L+1)} = W_{L+1} x_j^{(L)} + b_{L+1} \;\;\in\R^{Q}
\eeqn
contains no activation function, and that $\sigma$ has a Lipschitz continuous derivative (non-smooth activations like ReLU can be accommodated by a local mollification).

We let the vector of parameters $\uZ \in\R^K$ enlist all components of the weights $W_\ell$ and biases $b_\ell$, $\ell=1,\dots,L+1$, including those in the output layer, so that
\eqn
	K = \sum_{\ell=1}^{L+1} (M_\ell M_{\ell-1}+M_\ell )
\eeqn
where $M_0\equiv M$ accounts for the input layer.

In the output layer, we define  
\eqn
	x_j[\uZ] :=  x_j^{(L+1)}\in \R^Q
\eeqn 
for notational brevity, and obtain the $\cL^2$ cost as
\eqn\label{eq-cC-def-1-0}
	\cC[\ux[\uZ]] &=&\frac1{2N}\big|\ux[\uZ]-\uy_\omega\big|_{\R^{QN}}^2
	\nonumber\\
	&=& \frac1{2N}\sum_{j=1}^N |x_j[\uZ]-y_{\omega(j)}|_{\R^Q}^2 \,,
\eeqn 
where 
\eqn
	\ux := (x_1^T,\dots,x_N^T)^T\in\R^{QN} \,,
\eeqn 
and where $|\bullet|_{\R^n}$ denotes the Euclidean norm on $\R^n$.

\subsection{The standard gradient descent flow.}
The goal of the gradient descent algorithm is to find parameters $\uZ\in\R^K$ that minimize the cost function. The standard approach is based on the gradient flow  
\eqn\label{eq-uZ-graddesc-1-0}
	\partial_s \uZ(s) = -\nabla_{\uZ}\cC[\ux[\uZ(s)]] 
	\;\;,\;
	\uZ(0)=\uZ_0 \;\in\R^K \,.
\eeqn 
The vector field $\nabla_{\uZ}\cC[\ux[\bullet]]:\R^K\rightarrow\R^K $ is Lipschitz continuous if the derivative of the activation function $\sigma$ is Lipschitz. It then follows that the existence and uniqueness theorem for ordinary differential equations applies to \eqref{eq-uZ-graddesc-1-0}. 
From
\eqn\label{eq-cC-negder-1-0}
	\partial_s\cC[\ux[\uZ(s)]] 
	= - \big|\nabla_{\uZ}\cC[\ux[\uZ(s)]]\big|_{\R^K}^2
	\leq 0 \,,
\eeqn 
the cost $\cC[\ux[\uZ(s)]]$ is monotone decreasing in $s\in\R_+$, and since $\cC[\ux[\uZ(s)]]\geq0$ is bounded below, the limit $\cC_*=\lim_{s\rightarrow\infty}\cC[\ux[\uZ(s)]]$ exists for any orbit $\{\uZ(s)|s\in\R_+\}$, and depends on the initial data,
{\color{\redcol} $\cC_0=\cC[\ux[\uZ(0)]]$.} 
In applications, the initial data $\uZ_0\in\R^K$ is often chosen at random. 
 
The convergence of $\cC[\ux[\uZ(s)]]$ implies that $\lim_{s\rightarrow\infty}|\partial_s\cC[\ux[\uZ(s)]]|=0$, and hence, $\lim_{s\rightarrow\infty}|\nabla_{\uZ}\cC[\ux[\uZ(s)]]|_{\R^K}=0$ due to \eqref{eq-cC-negder-1-0}. Thus, the basic goal is to find $\cC_*=\lim_{s\rightarrow\infty}\cC[\ux[\uZ(s)]]=\cC[\ux[\uZ_*]]$ where $\uZ_*$ is a critical point of the gradient flow \eqref{eq-uZ-graddesc-1-0}, satisfying $0=-\nabla_{\uZ}\cC[\ux[\uZ_*]]$.

As discussed in \cite{cheewa-3}, there are major challenges to this strategy. First of all, the cost always converges to a stationary value, but not necessarily to the global minimum. Typically, there exist approximate or exact local minima (where the norm of the gradient is small or zero), and identifying ones that yield a sufficiently well-trained DL network is mostly based on ad hoc methods.

Moreover,  as $s\rightarrow\infty$, neither does $\lim_{s\rightarrow\infty}\cC[\ux[\uZ(s)]]=\cC[\ux[\uZ_*]]$ imply that $\uZ(s)$ converges to $\uZ_*$, nor to any other element of $\{\uZ_{**}\in\R^K\,|\,\cC[\ux[\uZ_{**}]]=\cC[\ux[\uZ_*]]\}$, nor that $\uZ(s)$ converges at all, without further assumptions on $\cC[\ux[\bullet]]$. Therefore, while  $\cC[\ux[\uZ(s)]]$ always converges to a stationary value of the cost function under the gradient descent flow \eqref{eq-uZ-graddesc-1-0}, $\uZ(s)$ cannot generally be assumed to converge to a minimizer $\uZ_*$. A simple illustrative example is given in \cite{cheewa-3}. Therefore, it is a key problem to be able to estimate $s_0>0$ such that $\uZ(s_0)$ determines a sufficiently well trained DL network, up to a prescribed precision.

As we will discuss in detail in the work at hand, a main source for these challenges is the definition of the gradient in \eqref{eq-uZ-graddesc-1-0} with respect to the Euclidean metric on $\R^K$; the choice of this particular Riemannian structure is ad hoc, and not motivated by the geometry of the problem. Choosing the Euclidean structure on the output layer of the DL network instead, we will obtain modified (preconditioned) gradient flows for $\uZ(s)$ both in the overparametrized and underparametrized cases that are better adapted to the inherent geometry of the problem. In the overparametrized case, we will show that the modified gradient flow always drives the $\cL^2$ cost to its global minimum, at a uniform exponential rate, provided that a rank condition holds.

\subsection{Euclidean gradient flow in the output layer.}
\label{ssec-gradxflow-1-0}
A basic fact that we will exploit extensively in this work is the circumstance that the cost \eqref{eq-cC-def-1-0} depends on the parameters $\uZ\in\R^K$ only via its dependence on $\ux[\uZ]\in\R^{QN}$, the vector of outputs from the last layer of the DL network. 

In preparation of the main subject of our discussion, we first introduce the following simple comparison model (which may at this point be considered as a priori unrelated to \eqref{eq-cC-negder-1-0}), defined by the gradient flow 
\eqn\label{eq-uxgrad-comp-1-0}
	\partial_s \ux(s) &=& - \nabla_{\ux}\cC[\ux(s)]
	\\
	\nonumber
	\ux(0)&=&\ux^{(0)} \in\R^{QN}
\eeqn
with $s\in\R_+$. The gradient here is defined with the Euclidean metric on $\R^{QN}$, and as will be explained in Section \ref{ssec-Geomadapt-1-0}, this system plays a major role in this work because the modified gradient flows that we will introduce on the space of parameters $\R^K$ are mapped to it (or a constrained version of it).
Component-wise, it corresponds to  
\eqn 
	\partial_s (x_j(s)-y_{\omega(j)})&=& - \frac1{N} (x_j(s)-y_{\omega(j)})
\eeqn 
for all $j=1,\dots,N$, and its explicit solution is given by
\eqn\label{eq-L2-convrate-1-0}
	x_j(s)-y_{\omega(j)} = e^{-\frac s{N}}(x_j(0)-y_{\omega(j)})
\eeqn 
with initial data $x_j(0)\in\R^{QN}$, so that using the expression \eqref{eq-cC-def-1-0}, we find
\eqn\label{eq-C-expon-1-0}
	\cC[\ux(s)] = e^{-\frac{2s}{N}} \cC[\ux(0)] \,.
\eeqn 
Notably, the exponential convergence rates in \eqref{eq-L2-convrate-1-0} and \eqref{eq-C-expon-1-0} are uniform with respect to the respective initial data.
From \eqref{eq-L2-convrate-1-0}, we obtain
\eqn	
	\ux_*:=\lim_{s\rightarrow\infty}\ux(s)= \uy_{\omega} \,.
\eeqn 
In particular, this is the unique global minimizer of the $\cL^2$ cost, since 
\eqn 
	\lim_{s\rightarrow\infty}\cC[\ux(s)]=\cC[\ux_*]= 0
\eeqn 
by continuity of $\cC$. Uniqueness follow from the fact that $\cC$ is strictly convex in the variable $\ux-\uy_\omega$.

\subsection{Geometrically adapted gradient descent}
\label{ssec-Geomadapt-1-0}

Perhaps surprisingly at first, the simplicity of \eqref{eq-uxgrad-comp-1-0} and explicitness of its solutions can be exploited for the DL network. Here, we present two geometrically adapted modifications of \eqref{eq-uZ-graddesc-1-0}, one for the underparametrized situation $K< QN$, and one for the overparametrized situation $K\geq QN$. Both will be naturally linked to the comparison model \eqref{eq-uxgrad-comp-1-0}.

The cost $\cC[\ux[\uZ]]$ depends on the weights and biases $\uZ\in\R^K$ only through its dependence on the vector $\ux[\uZ]\in\R^{QN}$ in the output layer, therefore defining
\eqn\label{eq-Dmat-def-1-0}
	D[\uZ]&:=&\Big[\frac{\partial x_j[\uZ]}{\partial \Z_\ell}\Big]_{j=1,\dots,N,\;\ell=1,\dots,K}
	\nonumber\\
	&=&
	\left[
	\begin{array}{ccc}
		\frac{\partial x_1[\uZ]}{\partial \Z_1} & \cdots & \frac{\partial x_1[\uZ]}{\partial \Z_K } \\
		\cdots & \cdots & \cdots \\
		\frac{\partial x_{N}[\uZ]}{\partial \Z_1} & \cdots & \frac{\partial x_{N}[\uZ]}{\partial \Z_K }
	\end{array}\right]
	\;\;\;
	\in \R^{QN\times K} \,,
\eeqn
we observe that 
\eqn\label{eq-nabZC-nabxC-1-0}
	\nabla_{\uZ}\cC[\ux[\uZ]] = D^T[\uZ]\nabla_{\ux}\cC[\ux[\uZ]] \,.
\eeqn 
Hence, \eqref{eq-uZ-graddesc-1-0} can be written in the form 
\eqn\label{eq-uZ-graddesc-1-0-1}
	\partial_s \uZ(s) = -D^T[\uZ(s)]\nabla_{\ux}\cC[\ux[\uZ(s)]] 
	\;\;,\;
	\uZ(0)=\uZ_0 \;\in\R^K \,,
\eeqn 
where, for the $\cL^2$ cost, we obtain the simple expression
\eqn 
	\nabla_{\ux}\cC[\ux[\uZ(s)]] = \frac1N(\ux[\uZ(s)]-\uy_\omega) \,.
\eeqn 
From here on, we will interchangeably write
\eqn 
	\ux(s) \equiv \ux[\uZ(s)] 
\eeqn 
for notational brevity.

The standard gradient flow \eqref{eq-uZ-graddesc-1-0-1} implies that
\eqn\label{eq-ux-graddesc-4-0-0}
	\partial_s \ux(s) &=& -D[\uZ(s)]D^T[\uZ(s)]\nabla_{\ux}\cC[\ux[\uZ(s)]]  
	\\
	\ux(0)&=&\ux[\uZ_0]\in\R^{QN}
	\nonumber
\eeqn  
The matrix $DD^T\in\R^{QN\times QN}$ determines the Neural Tangent Kernel (NTK) 
\eqn\label{eq-NTK-def-1-0}
	\Theta(x_i,x_j,\uZ)=(D[\uZ]D^T[\uZ])(i,j)
\eeqn 
via its $(i,j)$-th $Q\times Q$ block, for $i,j\in\{1,\dots,N\}$ \cite{jacgabhon}. In this sense, one may identify the NTK with $DD^T$.

We will discuss this fact in more detail in Section \ref{ssec-StdGradFlw-1-0}. The presence of $DD^T$ is the origin of the complicated cost landscape; 
{\color{\redcol}  it is known that if $DD^T$ is invertible, zero loss can be reached, \cite{lidingsun,zha}. However, } orbits can get trapped in approximate local minima (locations where the norm of the gradient is small) which typically occur in large quantities, and whose distribution is very difficult to control. We give a discussion of the invertibility of $DD^T$ in Section \ref{ssec-NTKrank-1-0}.

\subsubsection{The overparametrized case}
\label{sssec-overparam-1}
Instead of using the gradient flow corresponding to the Euclidean metric in $\R^K$, we will now introduce a modified gradient flow for the overparametrized DL network $K\geq QN$.

We assume that $D$ has full rank $QN$ in the region $U\subseteq\R^K$, and let
\eqn
	\Pen[D] = D^T(D D^T)^{-1} 
	\;\; \in \R^{K\times QN}
\eeqn 
denote the Penrose inverse of $D$ on $U$, also of rank $QN$. 
It generalizes the notion of the inverse of a matrix by way of
\eqn\label{eq-DPenD-id-1-0}
	D\Pen[D] = \1_{QN\times QN} \,,
\eeqn 
and 
\eqn
	P := \Pen[D] D = P^2 = P^T
\eeqn
is the projector onto the range of $D^T$, orthogonal with respect to the Euclidean metric on $\R^K$. 

Then, we define the modified (preconditioned) gradient flow on $U$,
\eqn\label{eq-uZ-graddesc-2-0-1}
	\partial_s \uZ(s) &=& -\Pen[D[\uZ(s)]](\Pen[D[\uZ(s)]])^T
	\nabla_{\uZ}\cC[\ux[\uZ(s)]]  \,.
\eeqn 
{\color{\redcol} We note that $P\,\Pen[D]=\Pen[D]D\Pen[D]=\Pen[D]$ due to \eqref{eq-DPenD-id-1-0}. Therefore, the right hand side of \eqref{eq-uZ-graddesc-2-0-1} is mapped by $P[\uZ(s)]$ to itself, and lies in the range of $P[\uZ(s)]$. Hence, \eqref{eq-uZ-graddesc-2-0-1} has the structure of a constrained dynamical system in $\R^K$. }

Explicitly, \eqref{eq-uZ-graddesc-2-0-1} can be written as 
\eqn\label{eq-uZ-graddesc-2-0}
	\partial_s \uZ(s) &=& -D^T[\uZ(s)](D[\uZ(s)]D^T[\uZ(s)])^{-2}D[\uZ(s)]
	\nabla_{\uZ}\cC[\ux[\uZ(s)]] 
	\nonumber\\
	\uZ(0)&=&\uZ_0 \;\in\R^K \,,
\eeqn 
and as will be shown in Theorem \ref{thm-overparam-1-0},  the vector $\ux(s)=\ux[\uZ(s)]\in\R^{QN}$ corresponding to the output layer of the DL network can be verified to satisfy
\eqn\label{eq-gradgcC-def-1-0}
	\partial_s\ux(s) = D[\uZ(s)]\partial_s\uZ(s)= -\nabla_{\ux}\cC[\ux(s)] \,,
\eeqn 
using \eqref{eq-uZ-graddesc-2-0} and \eqref{eq-nabZC-nabxC-1-0}.
That is, {\em it precisely corresponds to the Euclidean gradient flow of the comparison system 
{\color{\redcol} \eqref{eq-uxgrad-comp-1-0}},} 
and therefore exhibits the same uniform exponential convergence rates  as in \eqref{eq-L2-convrate-1-0} and \eqref{eq-C-expon-1-0}.

Therefore, \eqref{eq-uZ-graddesc-2-0-1} {\em always} drives the cost to the global minimum, for any initial data $\uZ(0)$ for which  $\rank(D[\uZ(s)])=QN$ is maximal along the associated orbit $\uZ(s)$. The latter condition ensures that $D[\uZ(s)]D^T[\uZ(s)]\in\R^{QN\times QN}$ is invertible, so that the right hand side of \eqref{eq-uZ-graddesc-2-0} is well-defined. In this situation,
\eqn
	\lim_{s\rightarrow\infty}\cC[\ux[\uZ(s)]] = 0
\eeqn 
holds 


The geometric structure will be described in detail in Section \ref{sec-overparam-1-0}; the map $\ux:\R^K\rightarrow\R^{QN}$ defines a horizontal vector bundle $\cV\subset T\R^K$, endowed with the pullback metric $h$ induced by the Euclidean structure on $\R^{QN}$. Then, the operator $\Pen[D](\Pen[D])^T\nabla_{\uZ}$ on the right hand side of  \eqref{eq-uZ-graddesc-2-0-1} is the coordinate representation of the gradient $\grad_h$ on $\cV$. This ensures that the vector field \eqref{eq-uZ-graddesc-2-0-1} is horizontal with respect to $\cV$, and in particular, that its pushforward under $\ux$ is the gradient vector field $-\nabla_{\ux}\cC[\ux]$ on $T\R^{QN}$  with the Euclidean metric. As we will explain in Section \ref{ssec-subRiem-1-0}, $\cV$ is in general non-holonomic, and the triple $(\R^K,\cV,h)$ defines a sub-Riemannian manifold.

In Section \ref{sec-rankloss-1-0}, we present a differential-algebraic system that generalizes \eqref{eq-uZ-graddesc-2-0} to allow for a loss of rank of $D$.

\subsubsection{The underparametrized case}
For the underparametrized DL network with $K< QN$, the map $\ux:\R^K\rightarrow\R^{QN}$, $\uZ\mapsto\ux[\uZ]$ is an embedding. We introduce the modified gradient flow defined by 
\eqn\label{eq-uZ-graddesc-3-0}
	\partial_s \uZ(s) &=& -(D^T[\uZ(s)]D[\uZ(s)])^{-1}\nabla_{\uZ}\cC[\ux[\uZ(s)]] 
	\nonumber\\
	\uZ(0)&=&\uZ_0 \;\in\R^K \,.
\eeqn
In a similar manner as in \eqref{eq-gradgcC-def-1-0}, one obtains that this is equivalent to the constrained dynamical system
\eqn 
	\partial_s\ux(s) = - \cP[\uZ(s)] \nabla_{\ux}\cC[\ux(s)] 
\eeqn 
for $\ux(s)=\ux[\uZ(s)]$, where
\eqn 
	\cP  = D(D^T D)^{-1}D^T =\cP^2  = \cP^T \,
\eeqn 
is the projector associated with the tangent bundle of the embedding $\ux:\R^K\rightarrow\R^{QN}$. 
This holds for any initial data $\uZ(0)$ for which  $\rank(D[\uZ(s)])=K$ is maximal along the associated orbit $\uZ(s)$. 
\\

\section{Geometrically adapted gradient flow for overparametrized DL}
\label{sec-overparam-1-0}

We will now give the precise statement of the results described in Section \ref{sssec-overparam-1}, for the overparametrized DL network with $K\geq QN$. We have the following main theorem.

\begin{theorem}\label{thm-overparam-1-0}
Given an overparametrized Deep Learning network with $K\geq QN$, assume for $D$ as in \eqref{eq-Dmat-def-1-0} that 
\eqn\label{eq-rankD-QN-1-0}
	\rank(D)=QN
\eeqn 
is maximal in the region $U\subset\R^K$. Let
\eqn\label{eq-Pen-def-1-0}
	\Pen[D] := D^T(D D^T)^{-1} \;\;\in\R^{K\times QN}
\eeqn 
denote the Penrose inverse of $D$ in $U$, with
\eqn\label{eq-P-def-1-0}
	\Pen[D]D = P
	\;\;,\;\; D\Pen[D] = \1_{QN\times QN} 
\eeqn 
where $P=P^2=P^T\in\R^{K\times K}$ is the orthoprojector onto the range of $D^T\in\R^{K\times QN}$.

Assume that $\uZ(s)\in U$, $s\in\R_+$, is a solution of the modified gradient flow
\eqn\label{eq-gradZ-overpar-1-0} 
	\partial_s\uZ(s) &=& - 
	\Pen[D[\uZ(s)]] (\Pen[D[\uZ(s)]] )^T\nabla_{\uZ}\cC[\ux[\uZ(s)]] 
	\nonumber\\
	\uZ(0)&=&\uZ_0 \in U\,.
\eeqn 
Then, the vector associated to the output layer $\ux(s)=\ux[\uZ(s)]\in\R^{QN}$ satisfies
\eqn 
	\partial_s\ux(s) = - \nabla_{\ux}\cC[\ux(s)]
	\;\;,\;\;
	\ux(0)=\ux[\uZ_0] \in \R^{QN} \,.
\eeqn 
In particular, for any arbitrary initial data $\uZ(0)=\uZ_0$ allowing for \eqref{eq-rankD-QN-1-0} to hold along the corresponding orbit $\uZ(s)$, $s\in\R_+$, the $\cL^2$ cost converges to its global minimum, 
\eqn\label{eq-cC-overparm-1-0} 
	\cC[\ux[\uZ(s)]]
	= e^{-\frac{2s}{N}}\cC[\ux[\uZ(0)]]
	\;\;\;(\rightarrow 0 \;{\rm as}\;s\rightarrow\infty)\,,
\eeqn 
and  
\eqn	
	\ux[\uZ(s)]-\uy_\omega = e^{-\frac{s}{N}}(\ux[\uZ(0)]-\uy_\omega)
	\;\;\;(\rightarrow 0 \;{\rm as}\; s\rightarrow\infty)\,,
\eeqn 
at the same uniform exponential convergence rates as in \eqref{eq-L2-convrate-1-0} and \eqref{eq-C-expon-1-0}.

If $\uZ(s)$ itself converges as $s\rightarrow\infty$, then $\uZ_*:=\lim_{s\rightarrow\infty}\uZ(s)$  satisfies
\eqn\label{eq-uZconv-nablC-1-0}
	\nabla_{\ux}\cC[\ux[\uZ_*]]
	= \frac1N(\ux[\uZ_*]- \uy_\omega)=0 \,,
\eeqn 
and $\uZ_*$ is a global minimizer, $\cC[\ux[\uZ_*]]=0$. 
\end{theorem}

\prf
Given that $\rank(D[\uZ])=QN$ is maximal for $\uZ\in U\subset\R^K$, 
it follows that $DD^T$ is invertible on $U$, so that \eqref{eq-Pen-def-1-0} and the right hand side of \eqref{eq-gradZ-overpar-1-0} are well-defined.
We have 
\eqn\label{eq-ux-ODE-1-0} 
	\partial_s\ux(s)
	 &=&D[\uZ(s)] \, \partial_{s}\uZ
	\nonumber\\
	&=&
	-  
	D[\uZ(s)]\underbrace{ D^T[\uZ(s)](D[\uZ(s)]D^T[\uZ(s)])^{-2} D[\uZ(s)]
	}_{=\Pen[D[\uZ(s)]] (\Pen[D[\uZ(s)]] )^T} 
	\nabla_{\uZ}\cC[\ux[\uZ(s)]]
	\nonumber\\
	&=&
	-  \underbrace{
	(D[\uZ(s)]D^T[\uZ(s)])^{-1}D[\uZ(s)] D^T[\uZ(s)]
	}_{=\1_{QN\times QN}} \nabla_{\ux}\cC[\ux[\uZ(s)]]
	\nonumber\\
	&=&
	-   \nabla_{\ux}\cC[\ux[\uZ(s)]]
	\nonumber\\
	&=&
	-   \nabla_{\ux}\cC[\ux(s)] \,.
\eeqn 
where we used \eqref{eq-nabZC-nabxC-1-0} to pass from the second to the third line.

Thus, it follows from the discussion in Section \ref{ssec-gradxflow-1-0} that if \eqref{eq-rankD-QN-1-0} is satisfied along the orbit $\uZ(s)$ for all $s\in\R_+$, then
\eqn
	\lim_{s\rightarrow\infty}\cC[\ux[\uZ(s)]] = 0 \,.
\eeqn 	
Convexity of $\cC[\ux]$ in $\ux-\uy_\omega$ implies that $\lim_{s\rightarrow\infty}\ux[\uZ(s)]=\uy_\omega$ converges. The convergence rates are as in \eqref{eq-L2-convrate-1-0} and \eqref{eq-C-expon-1-0}, because \eqref{eq-ux-ODE-1-0} is the same dynamical system as \eqref{eq-uxgrad-comp-1-0}.

If $\uZ_*=\lim_{s\rightarrow\infty}\uZ(s)$ exists, then 
\eqn
	\lim_{s\rightarrow\infty}\cC[\ux[\uZ[(s)]]] = \cC[\ux[\uZ_*]] = 0 \,,
\eeqn 
by continuity of $\cC$ and of $\ux[\bullet]$. Then, \eqref{eq-uZconv-nablC-1-0} follows from $\cC[\ux]=\frac N2|\nabla_{\ux}\cC[\ux]|^2$.
\endprf

\subsection{Rank condition on $DD^T$}
\label{ssec-NTKrank-1-0}

The rank of $DD^T$, which determines the NTK via \eqref{eq-NTK-def-1-0}, plays a major role in the above considerations. 
{\color{\redcol} In \cite{jacgabhon}, a detailed discussion is provided about the fact that in the infinite width limit of a deep network,  $DD^T$ is positive definite, and thus of full rank. Without the assumption of the infinite width limit, it is a subtle and challenging problem to determine sufficient conditions for full rank. In \cite{chemoo-1}, the NTK is proven to have full rank on an open subset of parameter space $\R^K$ in overparametrized deep networks with sufficient width and for generic training data; but the question whether this property is preserved along the entire gradient descent trajectory remains to be investigated in future work.  } Here, we present a few simple but insightful examples.

\subsubsection{Linear networks}
For a linear network determined by $f:\R^M\rightarrow\R^Q$, $f(x)=Wx+b$ with weight matrix $W\in\R^{Q\times M}$, bias $b\in\R^Q$, and training data $x_1,\dots,x_N\in\R^Q$, the Jacobian has the form
\eqn 
	D=\left.\left[
	\begin{array}{c}
		\1_{Q\times Q}\otimes x_1^T \\
		\cdots \\
		\1_{Q\times Q}\otimes x_N^T
	\end{array} 
	\right|
		u_{N}\otimes \1_{Q\times Q} 
	\right]
	\;\in\R^{NQ\times(QM+Q)}
\eeqn 
where the left block corresponds to derivatives in the row vectors of $W$, and the right block corresponds to the derivatives in the components of $b$. Here, 
\eqn
	\1_{Q\times Q}\otimes x^T=\left[
	\begin{array}{ccccc}
		x^T &0&\cdots&\cdots&0\\
		0&x^T&0&\cdots&0 \\
		\cdots&\cdots&\cdots&\cdots&\cdots\\
		0&\cdots&\cdots&0&x^T
	\end{array} 
	\right]
	\;\in\R^{Q\times QM}
\eeqn 
and we recall that $u_N=(1,\dots,1)^T\in\R^N$.
Assuming the overparametrized regime with $N\leq M$ and linear independence of $\{x_1,\dots,x_N\}$, the $NQ$ rows of $D$ are linearly independent, hence $\rank(D)=QN$, which implies that $DD^T$ is invertible. 

We also remark that in the underparametrized regime where $N>M$ and assuming that $\{x_1,\dots,x_N\}$ span $\R^M$, the rank of $D$ is maximally $QM<QN$. Hence, $DD^T$ is not invertible.

Next, we consider a two layer linear network, $f(x)=W_2(W_1 x + b_1)+b_2$ where we assume $x\in\R^M$, $b_1\in\R^{M_1}$, $b_2\in\R^{Q}$, and $W_1\in\R^{M_1\times M}$, $W_2\in\R^{Q\times M_1}$, with $M\geq M_1\geq Q$. Then, for training data $x_1,\dots,x_N\in\R^Q$, the Jacobian has the form 
\eqn\label{eq-D-twolin-1-0}
	D&=&\left.\left.\left.\left[
	\begin{array}{c}
		W_2\otimes x_1^T \\
		\cdots \\
		W_2\otimes x_N^T
	\end{array} 
	\right|
		u_{N}\otimes W_2
	\right|
	\begin{array}{c}
		\1_{Q\times Q}\otimes h_1^T \\
		\cdots \\
		\1_{Q\times Q}\otimes h_N^T
	\end{array} 
	\right|
		u_{N}\otimes \1_{Q\times Q} 
	\right]
	\nonumber\\
	&\in&
	\R^{NQ\times(QM+M_1+QM_1+Q)}
\eeqn 
where $h_j:=W_1 x_j + b_1$. We assume that either $N\leq M,M_1$ (overparametrized regime) and that $\{x_1,\dots,x_N\}$ are linearly independent. Then, if $W_2$ has full rank, then the rows of the $QN$ first block are linearly independent. If $W_1$ has full rank, then the $QN$ rows of the third block are linearly independent. In either case, $\rank(D)=QN$ is full, and $DD^T$ is invertible. 

However, if $\rank(W_1)=r_1<M_1$ and $\rank(W_2)=r_2<Q$, then the rank of $D$ is at most $r_1 N +r_2 Q$, which can be smaller than $NQ$. In the latter case, $DD^T$ is not invertible.

\subsubsection{Shallow network} Let us now consider a shallow network with ReLU activation $\sigma$, of the form $f(x)=W_2\sigma(W_1 x + b_1)+b_2$. We make the same assumptions on the weights and biases as for the two layer linear network.

If every ReLU is activated, then $\sigma(W_1 x_j + b_1)=h_j$, and we obtain the same expression for the Jacobian $D$ as in \eqref{eq-D-twolin-1-0}.
Assuming that $\{x_1,\dots,x_N\}$ are linearly independent, the rank of $D$ then depends on the ranks of $W_1$ and $W_2$ as given above.

Next, let us consider the example case where both $W_1$ and $W_2$ have full rank, but not all ReLU are activated. In particular, let us assume that in an open neighborhood of the given values of $W_i,b_i$, we have that $\sigma(W_1 x_j + b_1)=h_j$ are fully activated for $j=1,\dots,J$, but that $\sigma(W_1 x_j + b_1)=0$ are completely unactivated for $j=J+1,\dots,N$, after possibly relabeling the training data set.  Then, the Jacobian $D\in\R^{NQ\times(QM+M_1+QM_1+Q)}$ has the form
\eqn\label{eq-D-relu-1-0}
	D=\left.\left.\left.\left[
	\begin{array}{c}
		W_2\otimes x_1^T \\
		\cdots \\
		W_2\otimes x_J^T \\ 
		{\bf 0}_{Q(N-J)\times QM} 
	\end{array} 
	\right|
	\begin{array}{c}
		u_{J}\otimes W_2 \\ 
		{\bf 0}_{(N-J)Q\times M_1} 
	\end{array} 
	\right|
	\begin{array}{c}
		\1_{Q\times Q}\otimes h_1^T \\
		\cdots \\
		\1_{Q\times Q}\otimes h_J^T \\ 
		{\bf 0}_{Q(N-J)\times QM_1} 
	\end{array} 
	\right|
		u_{N}\otimes \1_{Q\times Q} 
	\right] 
	\;\;
\eeqn 
where ${\bf 0}_{m\times n}$ is the $m\times n$ matrix with all zero entries.

Given that $W_1$, $W_2$ have full rank, the first $J$ row vectors in the first and third block are linearly independent. 
Accordingly, $\rank(D)=JQ<NQ$ and $DD^T$ is not invertible.

If instead of ReLU, an activation function $\sigma:\R\rightarrow\R_+$ is used that is smooth and bijective (for instance softplus), the zero matrices as in \eqref{eq-D-relu-1-0} will not appear. However, due to its nonlinearity, $\sigma$ does not preserve linear independence; it may map a linearly dependent family of vectors to a linearly independent one, and vice versa. The determination of rank conditions for $D$ and control of invertibility of $DD^T$ is much more complicated in this case.

\subsection{A priori stopping condition}

While for the geometrically adapted gradient flow \eqref{eq-gradZ-overpar-1-0}, the $\cL^2$ cost always converges at an exponential rate to its global minimum, as long as the rank condition \eqref{eq-rankD-QN-1-0} holds, orbits $\uZ(s)$, $s\in\R_+$, do not necessarily converge. However, because of the exponential convergence of the cost along any orbit $\uZ(s)$ as given in \eqref{eq-C-expon-1-0}, we obtain the following sharp stopping time.

\begin{theorem}
Assume that $\rank(D D^T)=QN$ is maximal along the orbit $\uZ(s)$ of  \eqref{eq-gradZ-overpar-1-0}. Given $\epsilon>0$, let
\eqn
	s_{\epsilon,\uZ(0)} := \inf_{s\geq0}\big\{s\in\R_+\,\big|\,\cC[\ux[\uZ(s)]] \leq \epsilon\,\big\}
\eeqn
denote the earliest time at which the cost is $\leq\epsilon$. Then,
\eqn
	s_{\epsilon,\uZ(0)} =\frac N2\log\frac{\cC[\ux[\uZ(0)]]}\epsilon\,.
\eeqn 
\end{theorem}

\prf
We have
\eqn
	s_{\epsilon,\uZ(0)}  = \inf_{s\geq0}\big\{s\in\R_+\,\big|\,\cC[\ux[\uZ(s)]]=e^{-\frac{2s}{N}}\cC[\ux[\uZ(0)]] \leq \epsilon\,\big\} \,,
\eeqn
due to \eqref{eq-cC-overparm-1-0}. This immediately implies that
\eqn
	s_{\epsilon,\uZ(0)} =\frac N2\log\frac{\cC[\ux[\uZ(0)]]}\epsilon\,,
\eeqn 
as claimed.
\endprf

That is, for any given $\epsilon>0$, we obtain a precise stopping time for the modified gradient flow at 
\eqn
	s_{stop}\geq s_{\epsilon,\uZ(0)}  \,.
\eeqn 
The DL network trained with weights and biases $\uZ(s_{stop})$ then produces an $\cL^2$ cost $\cC[\ux[\uZ(s_{stop})]]\leq\epsilon$. 
The key advantage of this circumstance is that the minimum stopping time $s_{\epsilon,\uZ(0)} $ can be determined a priori; except for the initial value of the $\cL^2$ cost, it is independent of any orbit satisfying \eqref{eq-rankD-QN-1-0}. Therefore, a precise estimate on the training time based on the modified gradient descent flow can be given at the onset, provided that the rank condition $\rank(D)=QN$ in \eqref{eq-rankD-QN-1-0} is satisfied along the orbit.

\subsection{Gradient flow with respect to pullback bundle metric.}\label{ssec-geom-overpar-1-0}
We first describe the geometric setting for the general case, before specifying it for the situation in discussion. Let $\cM$ and $\cN$ be manifolds of dimensions $k=\dim(\cM)>n=\dim(\cN)$, and assume that $\cN$ is endowed with a Riemannian structure $(\cN,g)$ with metric $g$. Let  $f:\cM\rightarrow\cN$ be a smooth surjection, and let $Df:T\cM\rightarrow T\cN$ denote the associate differential map between the respective tangent bundles. Within this section, "smooth" will mean sufficiently regular, which will mostly be $C^1$. We denote the corresponding pullback map by $f^*$, and the pushforward map by $f_*$.

We denote by $\cV\subset T\cM$ a smooth horizontal vector bundle associated to the submersion $f:\cM\rightarrow\cN$, so that $ker(Df)\oplus \cV=T\cM$,  and $Df|_{\cV_x}:\cV_x\rightarrow T_{f(x)}\cN$ is a linear isomorphism at every $x\in\cM$. We note that $\cV$ can for instance be obtained by choice of an Ehresmann connection, or as the orthogonal complement of $ker(Df)\subset T\cM$ with respect to an auxiliary Riemannian structure on $\cM$. 

Let $\Gamma(\cM)$ and $\Gamma(\cV)$ denote the sets of smooth sections (vector fields) $\cM\rightarrow T\cM$ and $\cM\rightarrow\cV$, respectively. In particular, we may define the pullback metric $h$ on $\cV$ by way of
\eqn 
	h(V,W) = g(f_*V,f_*W)
\eeqn 
for sections $V,W\in\Gamma(\cV)$. Then, we can define the gradient $\grad_h$ associated to $(\cV,h)$ by way of
\eqn\label{eq-geominv-gradhF-1-0}
	d\cF(V) = h(V,\grad_h\cF)
\eeqn 
for all $V\in\Gamma(\cV)$, and any smooth $\cF:\cM\rightarrow\R$, where $d\cF$ denotes its exterior derivative. 

In our concrete situation, $\cM=\R^K$, $\cN\subseteq\R^{QN}$ open, so that $k=K$ and $n=QN$. Moreover, the coordinate representation of the map $f$ is $\ux:\R^K\rightarrow\R^{QN}$ and that of the pushforward $f_*$ applied to vector fields is the Jacobian matrix $[Df^\alpha_\beta]=D[\uZ]$ defined in \eqref{eq-Dmat-def-1-0}. For the pullback vector bundle $\cV$ under $\ux:\R^K\rightarrow\R^{QN}$, we have that the fiber $\cV_{\uZ}$ is given by the range of $D^T[\uZ]$. 
We claim that the modified gradient flow \eqref{eq-uZ-gradhC-1-0} is the gradient flow with respect to the bundle metric $h$ associated to $\cV$.

\begin{proposition}
Let $h$ denote the pullback metric on the vector bundle $\cV$ induced by the Euclidean metric on $T\R^{QN}$, as in \eqref{eq-hmetr-def-1-0}. Let $\grad_h$ denote the gradient with respect to $h$ as in \eqref{eq-gradh-def-1-0}. Then, writing $\widetilde\cC[\uZ]:=\cC[\ux[\uZ]]$ for the $\cL^2$ cost function, the geometrically invariant expression for \eqref{eq-gradZ-overpar-1-0} is given by
\eqn\label{eq-uZ-gradhC-1-0}
	\partial_s \uZ(s) = - \grad_h\widetilde\cC[\uZ(s)] \,.
\eeqn 
In particular, the gradient vector field in \eqref{eq-uZ-gradhC-1-0} is a section of $\cV$, and thus horizontal with respect to $\cV$.
\end{proposition}

\prf 
In local coordinates, we let $g_{\alpha,\alpha'}$ with $\alpha,\alpha'=1,\dots,n$ denote the matrix components of $g$, and $\sum_\beta Df^\alpha_\beta V^\beta$ with $\alpha=1,\dots,k$, $\beta=1,\dots,K$, the coordinate representation of $f_*V$ where $V= \sum_\beta V^\beta\partial_{\beta}$, and $Df^\alpha_\beta$ are the components of the Jacobian matrix of $f$. Then,  
\eqn\label{eq-hmetr-def-1-0}
	h(V,W)=\sum_{\alpha,\alpha',\beta,\beta'}
	g_{\alpha,\alpha'}Df^\alpha_\beta Df^{\alpha'}_{\beta'}V^\beta W^{\beta'} \,,
\eeqn 
so that \eqref{eq-geominv-gradhF-1-0} is equivalent to
\eqn\label{eq-grad-coord-1-0}
	\sum_\beta V^\beta\partial_\beta\cF = 
	\sum_{\alpha,\alpha',\beta,\beta'}
	g_{\alpha,\alpha'}Df^\alpha_\beta Df^{\alpha'}_{\beta'}V^\beta (\grad_h\cF)^{\beta'}
\eeqn 
for all $V\in\Gamma(\cV)$, from $d\cF(V)=\sum_\beta V^\beta\partial_\beta\cF$. 

The Riemannian structure on $\cN\subseteq\R^{QN}$ is given by the Euclidean metric, $g_{\alpha,\alpha'}=\delta_{\alpha,\alpha'}$ (the Kronecker delta) in Euclidean coordinates, so that \eqref{eq-grad-coord-1-0} has the form
\eqn\label{eq-gradFV-1-0}
	\uV^T\nabla_{\uZ}\cF = \uV^T D^T[\uZ] D[\uZ] \; \grad_h\cF \,,
\eeqn 
for all $\uV=(V^1,\dots,V^K)^T\in\ran(D^T[\uZ])$.  
Thus, this is equivalent to 
\eqn 
	P[\uZ]\nabla_{\uZ}\cF = P[\uZ] D^T[\uZ] D[\uZ] \; \grad_h\cF
\eeqn 
where $P=P^2=P^T\in\R^{K\times K}$ is the orthoprojector onto the range of $D^T$, as defined in \eqref{eq-P-def-1-0}. As noted above, the range of $D^T[\uZ]$ is, by construction, equivalent to the fiber $\cV_{\uZ}$. 
Applying the Penrose inverse of $D^T[\uZ]$ from the left, we obtain
\eqn 
	(\Pen[D[\uZ]])^T P[\uZ]\nabla_{\uZ}\cF
	= (\Pen[D[\uZ]])^T \nabla_{\uZ}\cF =   D[\uZ] \; \grad_h\cF \,,
\eeqn 
and subsequently applying the Penrose inverse of $D[\uZ] $ from the left, 
\eqn\label{eq-gradh-def-1-0}
	\grad_h\cF = \Pen[D[\uZ]](\Pen[D[\uZ]])^T  \nabla_{\uZ}\cF  \,.
\eeqn 
This is the expression for the gradient associated to the pullback metric $h$ on the vector bundle $\cV$, in local coordinates. Notably, $P\grad_h\cF =\grad_h\cF $, and $P^\perp\grad_h\cF =0$; that is, $\grad_h\cF \in\Gamma(\cV)$ as defined here is a section of $\cV$. This establishes the claim.
\endprf

\subsection{Relationship to sub-Riemannian geometry}\label{ssec-subRiem-1-0}

It is natural to ask whether the vector bundle $\cV$ is integrable (holonomic), i.e., whether $\R^K$ foliates into submanifolds whose tangent spaces coincide with the fibers of $\cV$. This holds if and only if the Frobenius condition is satisfied,
\eqn 
	\cV\;{\rm integrable} \; \Leftrightarrow \;
	[V,W]\in\Gamma(\cV) \;\;,\;\;\forall V,W\in\Gamma(\cV) \,,
\eeqn 
that is, the Lie bracket of any pair of sections of $\cV$ is a section of $\cV$.
To our knowledge, this is not generally the case for the situation under consideration.

To address the latter point, we observe that $\cV$ is the normal bundle to the foliation 
\eqn
	\R^K=\bigcup_{x\in\R^{QN}}f^{-1}(x)
	\;\;{\rm where}\;\;
	f:\R^{K}\rightarrow\R^{QN} \;,\;\uZ\mapsto\ux[\uZ] \,.
\eeqn 
That is, for any $\uZ\in f^{-1}(x)\subset\R^K$, $\cV_{\uZ}$ is normal to $T_{\uZ}f^{-1}(x)$. However, this does not imply integrability of $\cV$.
As an example of a foliation whose normal bundle is non-integrable, we may consider any smooth, non-integrable distribution $\widetilde\cV\subset T\R^K$ of fiber rank $K-1$. Then, $\widetilde{\cV}$ is normal to a vector bundle of fiber rank 1. The latter is  always integrable (the Frobenius condition is trivially satisfied, as the Lie bracket of any section with itself is zero), and thus generates a foliation of $\R^K$ into 1-dimensional curves. Accordingly, $\widetilde{\cV}$ is a non-integrable normal bundle to a foliation.

In case of $\cV$ being non-integrable, \eqref{eq-uZ-gradhC-1-0} is obtained by imposing non-holonomic constraints on the standard gradient flow \eqref{eq-uZ-graddesc-1-0}; the triple 
\eqn
	(\R^K, \cV, h)
\eeqn 
then defines a {\em sub-Riemannian manifold}. For some context on non-holonomic systems and sub-Riemannian geometry, we refer to \cite{arnnov-1,belris-1}.

\section{Comparison with the standard overparametrized gradient flow}
\label{ssec-StdGradFlw-1-0}
In comparison, we consider the standard gradient flow \eqref{eq-uZ-graddesc-1-0-1} with the Euclidean metric in the parameter space $\R^K$,
\eqn\label{eq-ux-graddesc-4-0}
	\partial_s \ux(s) &=& -D[\uZ(s)]D^T[\uZ(s)]\nabla_{\ux}\cC[\ux(s)] 
	\\
	\ux(0)&=&\ux[\uZ_0] \;\in\R^{QN} \,,
	\nonumber
\eeqn  
in the overparametrized scenario, $K>QN$, where $\ux(s)=\ux[\uZ(s)]$.
The operator $D D^T\nabla_{\ux}$ is the gradient on $T_{\ux}\R^{QN}$ associated to the metric tensor $(DD^T)^{-1}|_{f^{-1}(\ux)}$ (assuming invertibility) where $f:\R^K\rightarrow\R^{QN}$, $\uZ\mapsto\ux[\uZ]$.

\subsection{Trapping of orbits}
This metric is the source of the complicated cost landscape encountered for the standard gradient flow, and the emergence of a multitude of approximate local minima (where the norm of $DD^T\nabla_{\ux}\cC$ in \eqref{eq-ux-graddesc-4-0} is small) which can trap its orbits. 

{\color{\redcol}
We write the standard gradient descent flow \eqref{eq-uZ-graddesc-1-0} in the form 
\eqn\label{eq-uZ-graddesc-1-0-2}
	\partial_s \uZ(s) 
	&=&-D^T[\uZ(s)]\nabla_{\ux}\cC[\ux[\uZ(s)]] 
	\\
	\uZ(0)&=&\uZ_0 \;\in\R^K 
	\nonumber
\eeqn 
using \eqref{eq-nabZC-nabxC-1-0}. 

It is known that invertibility of $DD^T$ implies that zero loss can be reached, \cite{lidingsun,zha}. However, the}
main issue here arises from the fact that a priori, there exists no uniform lower bound on {\color{\redcol} } $DD^T$, that is, 
\eqn 
	\nexists \lambda>0 \,:\;\;
	(D D^T)[\uZ] > \lambda \;,\;\forall \uZ\in\R^K \,.
\eeqn 
Therefore, there is no mechanism precluding 
\eqn
	\inf_{\uZ\in\R^K}\inf_{v\in\R^{QN},|v|=1}
	|\, (D D^T)[\uZ] \, v \,|
\eeqn 
from being arbitrarily small, even if $DD^T>0$.
As a consequence, there may exist regions that "trap" orbits of \eqref{eq-uZ-graddesc-1-0-2} for a long time in the following sense, even if $\rank(D^T)=QN $ is maximal, so that $DD^T>0$ holds.


\begin{proposition}
\label{prp-trapping-1-0}
Assume that 
there exists a region $U\subset\R^K$ such that
\eqn
	0<|D^T[\uZ]\nabla_{\ux}\cC[\ux[\uZ]] |
	<\epsilon|\nabla_{\ux}\cC[\ux[\uZ]]|\ll1
	\;\;,\;\forall \uZ\in U \,,
\eeqn 
for some $0<\epsilon\ll1$. 
Moreover, assume that the orbit of the gradient flow \eqref{eq-uZ-graddesc-1-0-2} intersects $U$, that is, $\uZ(s)\in U$ for $s\in I\subseteq\R_+$, and let 
\eqn
	L_U := \big| \, \{\uZ(s)|s\in I\}\cap U \, \big| 
\eeqn
denote the arc length of its intersection with $U$.
Then, with $s_0:=\inf I$,
\eqn
	|I|\, \geq \,  \frac{N \, L_U}
	{|\ux[\uZ(s_0)]-\uy_\omega|}  \, \frac1{\epsilon}
\eeqn
is a lower bound on the time during which the orbit is trapped inside $U$.  
\end{proposition}

\prf
Clearly,  
\eqn 
	\nabla_{\ux}\cC[\ux] = \frac1N(\ux-\uy_\omega) \;\in\R^{QN} 
\eeqn 
is zero if and only if $\ux = \uy_\omega$, which globally minimizes the cost, due to \eqref{eq-cC-def-1-0}.

Then, we consider the arc length of the intersection of the orbit with $U$,
\eqn	
	L_U&=&\big| \, \{\uZ(s)|s\in\R_+\}\cap U \, \big| 
	\nonumber\\
	&=&  \int_I ds|\nabla_{\uZ}\cC[\ux[\uZ(s)]]|
	\nonumber\\
	&\leq&|I|\epsilon \; \sup_{s\in I}|\nabla_{\ux}\cC[\ux[\uZ(s)]]|
	\nonumber\\
	&=&
	|I|\epsilon \; \Big(\frac{2}{N} \sup_{s\in I}|\cC[\ux[\uZ(s)]]| \Big)^{\frac12}
	\nonumber\\
	&=&
	|I|\epsilon \; \Big(\frac{2}{N} |\cC[\ux[\uZ(s_0)]]| \Big)^{\frac12}
	\nonumber\\
	&=&
	\frac{|I|\epsilon}{N} \; |\ux[\uZ(s_0)]-\uy_\omega|
\eeqn 
where $s_0=\inf I$ is the initial time of entry of the orbit into $U$. Here, we used monotone decrease of the cost along the orbit to pass to the fifth line, and recalled \eqref{eq-cC-def-1-0} to pass to the last line. 
The claim follows.
\endprf

In the context of computationally determining orbits of the standard gradient descent flow, "trapping regions" of this type might turn out to be difficult to distinguish from true (degenerate) local extrema where $\nabla_{\uZ}\cC[\ux[\uZ]]=0$. Within the bounds of numerical precision for a given implementation, they might indeed be indistinguishable.

\subsection{Local extrema and rank loss}
In contrast to the situation addressed in Proposition \ref{prp-trapping-1-0} which assumes full rank of $D$, we note that true local extrema of \eqref{eq-uZ-graddesc-1-0-2} outside of the global cost minimum necessitate a rank loss, $\rank(D)<QN$, as shown in the following lemma.

\begin{lemma}
Any equilibrium solution $\uZ_*$ of the standard gradient flow \eqref{eq-uZ-graddesc-1-0-2} that is not a global  minimizer of the cost, i.e., $\nabla_{\uZ}\cC[\ux[\uZ_*]]=0$ but $\nabla_{\ux}\cC[\ux[\uZ_*]]\neq0$, must satisfy 
\eqn\label{eq-equil-rank-1-0}
	\rank(D[\uZ_*])<QN 
	\;\;{\rm and} \;\;
	\nabla_{\ux}\cC[\ux[\uZ_*]]\in \ker (D^T[\uZ_*])\,.
\eeqn 
\end{lemma}

\prf 
Let us first assume that $\rank(D[\uZ(s)])=QN$ along an orbit of \eqref{eq-uZ-graddesc-1-0-2}. If this orbit contains an isolated local equilibrium point $\uZ_*$ of \eqref{eq-uZ-graddesc-1-0-2}, then it must stem from 
\eqn\label{eq-nabcC-kerDT-1-0}
	0\neq \nabla_{\ux}\cC[\ux[\uZ_*]]\in \ker (D^T[\uZ_*]) \,.
\eeqn 
However, because the domain of $D^T[\uZ(s)])$ is $\R^{QN}$, the assumption $\rank(D[\uZ(s)])=QN$ implies that its kernel is trivial. Therefore, \eqref{eq-nabcC-kerDT-1-0} is not possible.

{\color{\redcol} Therefore, existence of an isolated local equilibrium point $\uZ_*$ of \eqref{eq-uZ-graddesc-1-0-2}, which is not a global minimum, requires that both $\rank(D[\uZ_*])<QN$ and \eqref{eq-nabcC-kerDT-1-0} hold. }
\endprf

\section{Rank loss in overparametrized Deep Learning networks}
\label{sec-rankloss-1-0}

In this section, we further investigate the standard and geometrically adapted DL networks in the overparametrized case, $K>QN$, but in the situation where a rank loss $\rank(D)<QN$ is allowed to occur.

\subsection{Rank loss and constrained dynamics}
If a rank loss occurs, $\rank(D^T)<QN$ in a set $U'\subset\R^{K}$, it follows that on $U'$, the system is effectively underparametrized. 
To elucidate this point, we diagonalize the positive semi-definite matrix
\eqn 
	DD^T = R^T\Lambda R \;\;,\;\uZ\in U'
\eeqn 
in $U'$, where $R:U'\rightarrow SO(QN)$, and $\Lambda$ maps $U'$ to the set of positive semi-definite diagonal matrices. Let $\widetilde{\Lambda^{-1}}$ denote its generalized inverse, so that $\cP_\Lambda :=\Lambda\widetilde{\Lambda^{-1}}=\widetilde{\Lambda^{-1}}\Lambda$ is the diagonal projector obtained from replacing all non-zero entries of $\Lambda$ by 1, while all other entries are zero. Accordingly, 
\eqn\label{eq-cP-rankloss-1-0}
	\cP := R^T \cP_\Lambda R \, = \,  \cP^2 \, = \, \cP^T
\eeqn  
is the orthoprojector (relative to the Euclidean metric on $\R^{QN}$) onto the range of $DD^T$ on $U'$. In particular, $DD^T=\cP DD^T \cP$ has full rank on the range of $\cP$, and  
\eqn
	(\cP D D^T \cP)^{-1}|_{\ran(\cP)} = R^T \widetilde{\Lambda^{-1}} R \,.
\eeqn
We also let $\cP^\perp:=\1_{QN\times QN}-\cP$.

\subsubsection{Rank loss in the standard gradient flow}
Given the above circumstances, the standard gradient descent flow yields 
\eqn\label{eq-standGF-ux-1-0}
	\partial_s \ux(s) &=& - \cP[\uZ(s)] D[\uZ(s)]D^T[\uZ(s)] \nabla_{\ux}\cC[\ux(s)]
	\,,
\eeqn 
for $\ux(s)=\ux[\uZ(s)]\in\R^{QN}$, $\uZ(s)\in U'$. 

Thus, it has the structure of a constrained dynamical system in $\R^{QN}$ relative to \eqref{eq-ux-graddesc-4-0}, due to the projection of the vector field to the range of $\cP$. This is analogous to the situation encountered in underparametrized DL networks addressed in \cite{cheewa-3}, and in the next section.  

\subsubsection{Rank loss in the geometrically adapted gradient flow}
The definition of the geometrically adapted gradient flow given in Theorem \ref{thm-overparam-1-0} needs to be modified in case a rank loss, $\rank(D)<QN$, is admitted. In fact, we instead define the following {\em differential-algebraic system}; for background on the efficient computational solution of differential-algebraic systems, see for instance \cite{hailub-1}.

\begin{theorem}\label{prp-overpar-rankloss-1-0}
Assume that $\rank(D)\leq QN$, and that the orbit $\uZ(s)$ solves the differential-algebraic system
\eqn\label{eq-uZ-diffalg-1-0}
	\partial_s\uZ(s) &=& D^T[\uZ(s)] \Psi[\uZ(s)]
	\\
	\Psi[\uZ(s)] &=&\argmin_\Psi\{\;|D[\uZ(s)]D^T[\uZ(s)]\Psi  + \nabla_{\ux}\cC[\ux[\uZ(s)]]\;|_{\R^{QN}}^2\;\}
	\nonumber\\
	\uZ(0)&=&\uZ_0\in\R^K \,.
	\nonumber
\eeqn 
That is, $\Psi[\uZ(s)]$ solves 
\eqn
	D[\uZ(s)]D^T[\uZ(s)]\Psi = - \nabla_{\ux}\cC[\ux[\uZ(s)]] 
	\;\;+ \; {\rm minimal\; error\; in}\;L^2\,
\eeqn
via least square optimization. 

Then, if $rank(D)=QN$ is maximal, \eqref{eq-uZ-diffalg-1-0} is equivalent to the geometrically adapted gradient flow in  Theorem \ref{thm-overparam-1-0}.

If $\rank(D)<QN$, then  $\ux(s)=\ux[\uZ(s)]$ solves
\eqn\label{eq-uxdyn-diffalg-1-0}
	\partial_s\ux(s) &=&
	-\cP[\uZ(s)]\nabla_{\ux}\cC[\ux[\uZ(s)]]
	\\
	\nonumber
	\ux(0)&=&\ux[\uZ_0] \in \R^{QN} \,,
\eeqn
where the projector $\cP$ onto the range of $DD^T$ is defined in \eqref{eq-cP-rankloss-1-0}.
\end{theorem}

\prf
Evidently, if $\rank(D)=QN$ is maximal, then $DD^T$ is invertible, and \eqref{eq-uZ-diffalg-1-0} reduces to the definition in  Theorem \ref{thm-overparam-1-0}. 
On the other hand, if $\rank(D)<QN$, then it follows from 
\eqn 
	\lefteqn{
	|D[\uZ(s)]D^T[\uZ(s)]\Psi  + \nabla_{\ux}\cC[\ux[\uZ(s)]]\;|_{\R^{QN}}^2
	}
	\\
	&=&
	|\cP[\uZ(s)] D[\uZ(s)]D^T[\uZ(s)]\Psi  + \cP[\uZ(s)]\nabla_{\ux}\cC[\ux[\uZ(s)]]\;|_{\R^{QN}}^2
	\nonumber\\
	&&\hspace{4cm}
	+
	|\cP^\perp[\uZ(s)]\nabla_{\ux}\cC[\ux[\uZ(s)]]\;|_{\R^{QN}}^2
	\nonumber
\eeqn 
that the right hand side is minimized when the first term on the rhs is zero, whence
\eqn
	(\cP D D^T \cP)[\uZ(s)]  \Psi[\uZ(s)]  = - \cP[\uZ(s)]\nabla_{\ux}\cC[\ux[\uZ(s)]]
\eeqn
where we used that $\cP D D^T=DD^T=D D^T\cP$. Because $\cP DD^T \cP$ has full rank on the range of $\cP$, we find that 
\eqn
	 \Psi[\uZ(s)]  = -(\cP D D^T \cP)^{-1}|_{\ran(\cP)}[\uZ(s)] \cP[\uZ(s)]\nabla_{\ux}\cC[\ux[\uZ(s)]]
\eeqn
We thus find for $\ux(s)=\ux[\uZ(s)]$ that
\eqn
	\partial_s\ux(s) &=& D[\uZ(s)]\partial_s\uZ(s)
	\nonumber\\
	&=&( DD^T)[\uZ(s)] \Psi[\uZ(s)]
	\nonumber\\
	&=&
	-\big( \, (\cP DD^T \cP)(\cP D D^T \cP)^{-1}|_{\ran(\cP)} \, \big)[\uZ(s)] \cP[\uZ(s)]\nabla_{\ux}\cC[\ux[\uZ(s)]]
	\nonumber\\
	&=&
	-\cP[\uZ(s)]\nabla_{\ux}\cC[\ux[\uZ(s)]] \,.
\eeqn
This proves the claim.
\endprf

\begin{theorem}
A point  $\uZ_*\in\R^K$ is an equilibrium of the standard gradient flow \eqref{eq-uZ-graddesc-1-0-2} if and only if it is an equilibrium either of the geometrically adapted gradient flow \eqref{eq-uZ-diffalg-1-0} if $D$ does not have full rank, {\color{\redcol}  or of \eqref{eq-gradZ-overpar-1-0} if $D$ has full rank.}

Assume that the activation function $\sigma$ is smooth.
If $\rank(\cP)=r<QN$ in an open neighborhood $U\subset\R^K$, then local equilibria of \eqref{eq-uZ-graddesc-1-0-2} and \eqref{eq-uZ-diffalg-1-0} which are not global are generically contained in an $(K-r)$-dimensional critical submanifold $\cM_{crit}\subset U$, in the sense of Sard.  
\end{theorem}

\prf
If a rank loss occurs, $\rank(D^T)<QN$, in a neighborhood $U'\subset\R^{K}$, then an equilibrium of \eqref{eq-standGF-ux-1-0} and an equilibrium of \eqref{eq-uxdyn-diffalg-1-0} both require the same condition  
\eqn 
	0 = - \cP[\uZ_*] \nabla_{\ux} \cC[\ux[\uZ_*]]
	\;\;,\;\uZ_*\in U'\,.
\eeqn
For \eqref{eq-standGF-ux-1-0}, this is because of $ DD^T = \cP DD^T \cP$ has full rank on the range of $\cP$. Therefore, any equilibrium of  \eqref{eq-gradZ-overpar-1-0} that is not a global minimizer satisfies \eqref{eq-equil-rank-1-0}. 

Next, assume that $\rank(\cP)=r<QN$ on an open neighborhood $U\subset\R^K$, and let $\uZ\in U$. Given that the activation function $\sigma$ is smooth, both $D[\uZ]$ and $\ux[\uZ]$ depend smoothly on $\uZ$. Assume that the linearly independent family of smooth vector fields $V_\alpha:\R^K\rightarrow\R^{QN}$, $\alpha=1,\dots,r$, spans the range of $\cP$; for instance, one could choose a suitable $r$-tuple of column vectors of $D$. 
Then, we define the smooth family of functions $g_\alpha:\R^K\rightarrow \R$ obtained from the inner product
\eqn\label{eq-galpha-def-1-0}
	g_\alpha[\uZ]&:=&\big(\,V_\alpha[\uZ]\,,\,\cP[\uZ] \nabla_{\ux} \cC[\ux[\uZ]]\,\big)_{\R^{QN}}
	\nonumber\\
	&=&\big(\,V_\alpha[\uZ]\,,\, \nabla_{\ux} \cC[\ux[\uZ]]\,\big)_{\R^{QN}}
	\;,\;\alpha=1,\dots,r \,.
\eeqn 
The set of equilibrium solutions in $U\subset\R^K$ is given by
\eqn 
	\cM_{crit}=U\cap\bigcap_{\alpha=1}^r g_\alpha^{-1}(0) \,.
\eeqn 
Then, by Sard's theorem, $\cM_{crit}\subset U\subset\R^K$ is generically a $(K-r)$-dimensional submanifold.
\endprf

We conclude that the standard gradient flow \eqref{eq-uZ-graddesc-1-0-2} and the geometrically adapted gradient flow \eqref{eq-gradZ-overpar-1-0} have the same set of equilibria. If those equilibria are not global, then they are generically not isolated points, but elements of a critical manifold. On the other hand,  \eqref{eq-gradZ-overpar-1-0} exhibits a uniform exponential convergence rate whenever $DD^T$ has full rank, as opposed to  \eqref{eq-uZ-graddesc-1-0-2} which provides no such a priori control. 
{\color{\redcol} We also note that in \cite{coo}, it is shown that global minima in overparametrized networks define critical submanifolds. }

\subsection{Perturbation away from local equilibrium}
If an orbit converges towards (is trapped by) a local equilibrium $\uZ_*\in \cM_{crit}$, one may wish to select an updated initial condition $\uZ_0'$ to continue the gradient descent either based on the standard or geometrically adapted flow, that is, \eqref{eq-standGF-ux-1-0} or \eqref{eq-uxdyn-diffalg-1-0}, respectively. A suitable choice is obtained from
\eqn
	\uZ_0'=\uZ_*+ V_{\uZ_*}
\eeqn
where $\ux_*\in \cM_{crit}$, and $V_\bullet:\cM_{crit}\rightarrow N\cM_{crit}$ is a section of the normal bundle of $\cM_{crit}$. Since $N\cM_{crit}$ is spanned by $\{\nabla_{\uZ}g_\alpha\big|_{\cM_{crit}}\}_{\alpha=1}^r$, one can for example pick
\eqn\label{eq-normalV-1-0}
	V_{\uZ_*} = \kappa_\alpha  \nabla_{\uZ} g_\alpha\big|_{\uZ=\uZ_*}
\eeqn 
with $g_\alpha$ defined as in \eqref{eq-galpha-def-1-0}, and a suitable choice of $\alpha\in\{1,\dots,r\}$.
The sign of the parameter $\kappa_\alpha\in\R$ is chosen such that the cost is decreasing,
\eqn\label{eq-cC-monot-1-0}
	\cC[\ux[\uZ_0']] < \cC[\ux[\uZ_*]] \,.
\eeqn
More generally, one may choose a linear superposition of vectors of the form \eqref{eq-normalV-1-0}, for each of which one obtains \eqref{eq-cC-monot-1-0}.

\section{Geometrically adapted gradient flow for underparametrized DL}

In the underparametrized situation $K<QN$, we obtain the following theorem.

\begin{theorem}\label{thm-underparam-1-0}
Assume that $K<QN$, and that 
\eqn\label{eq-rankD-K-1-0}
	\rank(D[\uZ])=K
\eeqn 
is maximal in the region $\uZ\in U\subset\R^K$. 
Assume that $\uZ(s)\in U$, $s\in [0,T)\subseteq\R_+$, is an orbit of the modified gradient descent flow  
\eqn\label{eq-uZ-graddesc-2-1}
	\partial_s \uZ(s) &=& -(D^T[\uZ(s)]D[\uZ(s)])^{-1}\nabla_{\uZ}\cC[\ux[\uZ(s)]] 
	\\
	\uZ(0)&=&\uZ_0 \;\in U\,.
	\nonumber
\eeqn
Then, $\ux(s)=\ux[\uZ(s)]]$ is a solution to the constrained gradient flow defined by 
\eqn\label{eq-uxdyn-underpar-1-0}
	\partial_s\ux(s) &=& - \cP[\uZ(s)] \nabla_{\ux}\cC[\ux[\uZ(s)]] 
	\\
	\nonumber
	\ux(0)&=&\ux[\uZ_0] \in\R^{QN}
\eeqn 
where
\eqn\label{eq-def-cP-1-0} 
	\cP  = D(D^T D)^{-1}D^T \;\;\in\R^{QN\times QN}
\eeqn 
is the projector onto the range of $D$, orthogonal with respect to the Euclidean metric on $\R^{QN}$, satisfying $\cP=\cP^2=\cP^T$. In particular,
\eqn 
	\cP^\perp[\uZ(s)] \partial_s\ux(s) = 0
\eeqn 
where $\cP^\perp =\1_{QN\times QN}-\cP$.

Moreover, any critical point $\uZ_*\in U$ satisfies
\eqn\label{eq-cPnabC-1-0}
	0 = \cP[\uZ_*]\nabla_{\ux}\cC[\ux[\uZ_*]] \,,
\eeqn 
and the local extremum of the cost function at $\uZ_*$ is attained at
\eqn\label{eq-cCmin-GD-1-0}
	\cC[\ux[\uZ_*]] = \frac N2 
	\big|\cP^\perp[\uZ_*]\nabla_{\ux}\cC[\ux[\uZ_*]] \big|_{\R^{QN}}^2 
\eeqn 
where $\rank(\cP^\perp[\uZ_*])= QN-K$. 
\end{theorem}

\prf
Given that $\rank(D D^T)=K$ is maximal in $U$, it follows that the matrix $D^T D\in\R^{K\times K}$ is invertible. Therefore, the expression for the right hand side of  \eqref{eq-uZ-graddesc-2-1}, and \eqref{eq-def-cP-1-0} for the orthoprojector $\cP$, are well-defined. 

We have 
\eqn\label{eq-ux-ODE-underpar-1-0} 
	\partial_s\ux(s)
	 &=&D[\uZ(s)] \, \partial_{s}\uZ(s)
	 \nonumber\\
	&=&
	-   
	D[\uZ(s)] (D^T[\uZ(s)]D[\uZ(s)])^{-1} 	 	
	\nabla_{\uZ}\cC[\ux[\uZ(s)]]
	\nonumber\\
	&=&
	-  
	\underbrace{
	D[\uZ(s)] (D^T[\uZ(s)]D[\uZ(s)])^{-1} D^T[\uZ(s)]
	}_{=\cP[\uZ(s)] } 
	\nabla_{\ux}\cC[\ux[\uZ(s)]]
	\nonumber\\
	&=&
	-  \cP[\uZ(s)] \nabla_{\ux}\cC[\ux[\uZ(s)]]
	\nonumber\\
	&=&
	-  \cP[\uZ(s)]\nabla_{\ux}\cC[\ux(s)] \,,
\eeqn 
where we used \eqref{eq-uZ-graddesc-2-1} to pass to the second line.
As a consequence,
\eqn 
	\cP^\perp[\uZ(s)]\partial_s\ux(s) = 0 \,.
\eeqn 
It follows from the first line in \eqref{eq-ux-ODE-underpar-1-0} that $\partial_s \uZ(s)=0$ implies $\partial_s\ux(s)=0$. 

Let $\uZ_*$ denote a stationary point for \eqref{eq-uZ-graddesc-2-1}. Then, \eqref{eq-ux-ODE-underpar-1-0} implies that
\eqn 
	\cP[\uZ_*]\nabla_{\ux}\cC[\ux[\uZ_*]] = 0 \,.
\eeqn  
Therefore, 
\eqn
	\cC[\uZ_*] &=& \frac N2 |\nabla_{\ux}\cC[\ux[\uZ_*]] |^2
	\nonumber\\
	&=&
	\frac N2 \Big(\big|\cP[\uZ_*]\nabla_{\ux}\cC[\ux[\uZ_*]] \big|_{\R^{QN}}^2
	+ \big|\cP^\perp[\uZ_*]\nabla_{\ux}\cC[\ux[\uZ_*]] \big|_{\R^{QN}}^2 \Big)
	\nonumber\\
	&=&
	\frac N2 \big|\cP^\perp[\uZ_*]\nabla_{\ux}\cC[\ux[\uZ_*]] \big|_{\R^{QN}}^2
\eeqn  
as claimed, where $\rank(\cP^\perp[\uZ_*])= QN-K$.
\endprf

\begin{remark}
We note that along every orbit $\uZ(s)$ of \eqref{eq-uZ-graddesc-2-1},
\eqn 
	\partial_s\cC[\ux[\uZ(s)]] = - \big|\cP[\uZ(s)]\nabla_{\ux}\cC[\ux[\uZ(s)]] \big|^2
	\leq 0 \,,
\eeqn 
hence the $\cL^2$ cost is monotone decreasing. However, \eqref{eq-cCmin-GD-1-0} is generically strictly positive, and the modified gradient descent does not produce a global minimum of the cost. This matches the results for the standard gradient descent in the underparametrized case obtained in \cite{cheewa-3}.
\end{remark}

\subsection{Induced gradient flow in output layer}


Let $f:\R^K\rightarrow\R^{QN}$ denote the map locally represented by $\ux:\uZ\mapsto \ux[\uZ]$, so that 
\eqn\label{eq-cM-underpar-1-0}
	f(\R^K)=\cM\subset\R^{QN}
\eeqn
For convenience of presentation, let us here assume that $\cM$ is an embedded submanifold with constant dimension $\dim(\cM)=K$ everywhere.
Then, the tangent subbundle $T\cM\subset \cup_{p\in\cM}T_p\R^{QN}$ is 
spanned by sections $f_*V\in \Gamma(T\cM)$ for sections $V\in \Gamma(T\R^K)$. In local coordinates, $f_*$ is represented by the Jacobian matrix $Df^\alpha_\beta$ when acting on vector fields, with $\alpha=1,\dots,QN$ and $\beta=1,\dots,K$.

The Euclidean metric on the tangent bundle $T\R^{QN}$ induces a Riemannian metric on $T\cM$, which in turn defines the pullback metric $g$ on $T\R^K$ given by
\eqn\label{eq-hmetr-def-1-1}
	g(V,W)=\sum_{\alpha,\alpha',\beta,\beta'}
	\delta_{\alpha,\alpha'}Df^\alpha_\beta Df^{\alpha'}_{\beta'}V^\beta W^{\beta'} \;\;,\;\;
	V,W\in\Gamma(\R^K) \,,
\eeqn 
with $\alpha,\alpha'=1,\dots,QN$ and $\beta,\beta'=1,\dots,K$, in local coordinates where the Euclidean metric has components $\delta_{\alpha,\alpha'}$.  This is similar as in \eqref{eq-hmetr-def-1-0}. 
With $[Df^\alpha_\beta]=D[\uZ]$, as defined in \eqref{eq-Dmat-def-1-0}, we therefore obtain
\eqn 
	g_{\beta,\beta'}[\uZ] = (D^T[\uZ]D[\uZ])_{\beta,\beta'} \,,
\eeqn  
which endows $\R^K$ with the pullback Riemannian structure $(\R^K,g)$.

The corresponding gradient operator, defined by $d\cF[V]=g(V,\grad_g \cF)$ for any smooth vector field $V:\R^{K}\rightarrow T\R^{K}$ and smooth function $\cF:\R^{K}\rightarrow\R$ (where $d$ denotes the exterior derivative), has the form
\eqn\label{eq-grad-coord-1-0}
	\sum_{\beta} V^\beta\partial_\beta\cF = 
	\sum_{\beta,\beta'}
	g_{\beta,\beta'}
	V^\beta (\grad_g\cF)^{\beta'}
\eeqn 
in local coordinates. Since this holds for all $\uV=(V^1,\dots,V^K)^T\in\R^K$, we conclude that 
\eqn 
	\grad_g\cF &=& g^{-1}[\uZ]\nabla_{\uZ}\cF
	\nonumber\\
	&=& (D^T[\uZ]D[\uZ])^{-1}\nabla_{\uZ}\cF
\eeqn 	
in local coordinates $\uZ\in U_\alpha$. We obtain the following result.

\begin{proposition}
The coordinate invariant expression for the gradient flow generated by $\widetilde\cC[\uZ]:=\cC[\ux[\uZ]]$ on $(\R^K,g)$ is given by
\eqn\label{eq-uZgradgflow-1-0}
	\partial_s\uZ(s) = -\grad_g\widetilde\cC[\uZ(s)] \,.
\eeqn 
The modified gradient flow \eqref{eq-uZ-graddesc-2-1} is the local coordinate representation of \eqref{eq-uZgradgflow-1-0}. 
\end{proposition}

In contrast, the standard gradient flow \eqref{eq-uZ-graddesc-1-0} is defined with respect to the Euclidean metric in $\R^K$.

\begin{remark}
For both the underparametrized case, and the overparametrized case with rank loss, one obtains the constrained Euclidean gradient flow in the output layer of the form \eqref{eq-uxdyn-underpar-1-0}. In the underparametrized case addressed in Theorem \ref{thm-underparam-1-0}, the projector $\cP$ is associated to the tangent bundle of the submanifold $\cM$ in \eqref{eq-cM-underpar-1-0}, and therefore,  \eqref{eq-uxdyn-underpar-1-0} has the structure of a dynamical system with holonomic contraints. In the overparametrized case with rank loss addressed in Theorem \ref{prp-overpar-rankloss-1-0}, the projector $\cP$ is not in general affiliated with a foliation of $\R^{QN}$, and therefore, \eqref{eq-uxdyn-diffalg-1-0} has the structure of a dynamical system with non-holonomic constraints.
\end{remark}

\section{The borderline case $K=QN$}
\label{sec-borderline-1-0}

Finally, we remark that in the borderline case $K=QN$ separating the strictly over- and underparametrized situations, we have that $D\in\R^{K\times K}$ is a square matrix, and the condition that $\rank(D)=K$ is maximal implies that it is invertible.
Then, the Penrose inverse is equal to the ordinary inverse, 
\eqn 
	\Pen[D] = D^{-1}
\eeqn  
and we find that the modified gradient flow \eqref{eq-gradZ-overpar-1-0} for the overparametrized system attains the form
\eqn\label{eq-gradZ-overpar-1-2} 
	\partial_s\uZ(s) &=& - 
	D^{-1}[\uZ(s)]D^{-T}[\uZ(s)]\nabla_{\uZ}\cC[\ux[\uZ(s)]]
	\nonumber\\
	\uZ(0) &=& \uZ_0 \in \R^K \,.
\eeqn 
On the other hand, the expression for the gradient vector field for the underparametrized system \eqref{eq-uZ-graddesc-2-1} can be simplified by use of
\eqn 
	(D^T D)^{-1} = D^{-1} D^{-T} \,,
\eeqn
and thereby reduces to the same expression \eqref{eq-gradZ-overpar-1-2}.

Therefore, we conclude that the expressions for the modified gradient flows for the over- and underparametrized systems coincide in the borderline case $K=QN$.
In particular, $\ux(s)=\ux[\uZ(s)] $ then satisfies
\eqn 
	\partial_s\ux(s) = -\nabla_{\ux}\cC[\ux(s)] \,
\eeqn 
as in the overparametrized case. In particular, this is equivalent to the comparison system given in \eqref{eq-uZ-graddesc-1-0}, corresponding to the gradient flow in the output layer relative to the Riemannian structure determined by the Euclidean metric.
\\

\noindent
{\bf Acknowledgments:} 
The author thanks Igor Zelenko and Patricia Mu\~{n}oz Ewald for enlightening discussions. He thanks the anonymous referees for helpful comments, and gratefully acknowledges support by the NSF through the grant DMS-2009800, and the RTG Grant DMS-1840314 - {\em Analysis of PDE}. 

There are no financial or non-financial competing interests that are directly or indirectly related to this work.
\\

\end{document}